# Fingers' Angle Calculation using Level-Set Method

### Ankit Chaudhary
Dept. of Electrical & Computer Engg., The University of Iowa, USA
ankit-chaudhary@uiowa.edu

### Jagdish Lal Raheja
Machine Vision Lab, CEERI/CSIR, Pilani, INDIA
jagdish@ceeri.ernet.in

### Karen Das
Dept. of Electronics & Communication Engg., Don Bosco University, INDIA
karenkdas@gmail.com

### Shekhar Raheja
Dept. of Computer Science, TU Kaiserslautern, GERMANY
shekhar.raheja@gmail.com

*Abstract*—**In the current age, use of natural communication in human computer interaction is a known and well installed thought. Hand gesture recognition and gesture-based applications has gained a significant amount of popularity amongst people all over the world. It has a number of applications ranging from security to entertainment. These applications generally are real time applications and need fast, accurate communication with machines. On the other end, gesture based communications have few limitations also like bent finger information is not provided in vision based techniques. In this paper, a novel method for fingertip detection and for angle calculation of both hands' bent fingers is discussed. Angle calculation has been done before with sensor based gloves/devices. This study has been conducted in the context of natural computing for calculating angles without using any wired equipment, colors, marker or any device. The pre-processing and segmentation of the region of interest is performed in a HSV color space and a binary format respectively. Fingertips are detected using level-set method and angles were calculated using geometrical analysis. This technique requires no training for system to perform the task.**

## I. INTRODUCTION

Robust and natural hand gesture recognition from video or in real time is one of the most important challenges for researchers working in the area of computer vision. Gesture recognition systems are very helpful in general purpose life as they can be used by general people without any training as everybody know how to use hand and what sign would make what mean. So, if computers can understand gestures efficiently, computers would be more useful for all. It can also help in controlling devices, interacting with machine interfaces, monitoring human activities and in many other applications. Generally defined as any meaningful body motion, gestures play a central role in everyday communication and often convey emotional information about the gesticulating person. There are some specific gestures which are pre-defined in a particular community or society as sign language, but many gestures made by hand are just a random shape. Precisely all shapes made by hand gesture are not defined, so one need to track all shapes to efficiently control machines by hand gesture.

During the last few decades researchers have been interested in recognizing automatically human gestures for several applications like sign language recognition, socially assistive robotics, directional indication through pointing, control through gestures, alternative computer interfaces, immersive game technology, virtual controllers, affective computing and remote controlling. For further details on gesture applications see [Chaudhary et al., 2011][Mitra & Acharya, 2007]. Mobile companies are also trying to make handsets which can recognize gestures and operate over small distances [Kroeker, 2010][Tarrataca, Santos, & Cardoso, 2009]. There have been many non-natural methods using devices and color papers/rings. In the past, researchers have employed gloves [Sturman, & Zeltzer, 1994], color strips [Do et al., 2006][Premaratne, & Nguyen, 2007][Kohler, 1996][Bretzner et al., 2001] or full sleeve shirt [Kim & Fellner, 2004][Sawah et al., 2007] in image processing based methods to obtain better segmentation results. A preliminary part of this work has been published in [Chaudhary et al., 2012].

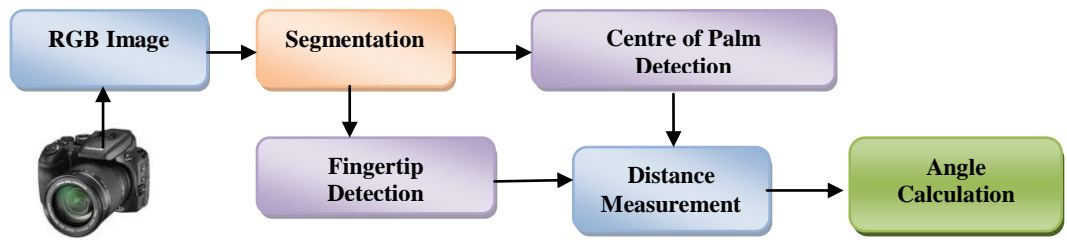

Figure 1: Algorithmic flow for angle approximation for both hands.

It was a well-known fact in advance that natural computing methods will take over other technologies. Pickering [2005] stated "initially touch-based gesture interfaces would be popular, but non-contact gesture recognition technologies would be more attractive finally". Recently human gesture recognition catches the peak attention of the research in both software and hardware environments. Many mobile companies like Samsung, Micromax have implemented hand gesture as a way to control mobile applications, which make it more popular in public domain. It can be used for controlling a robotic hand which can mimic the human hand actions and can secure human life by being used in many commercial and military operations. One such robotic hand is Dexterous from Shadow Robot®.

Many mechanical [Huber & Grupen, 2002][Lim et al., 2000] and image processing [Nolker & Ritter, 2002] based techniques are available in the literature to interpret single hand gesture. However, in generic scenario humans express their actions with both hands along. it is a new challenge to take into account the human hand actions depicted by both hands simultaneously. The computational time for both hand gestures would be more compared to that required for a single hand. The approach employed for a single hand can also be used for this purpose with a slight modification in the algorithm for both hands. However, the process may consumes twice time that is required for single hand gesture recognition in serial implementation.

There are certain conditions from [Nolker & Ritter, 2002] . If this algorithm is applied on both hands, it would not always take double time to calculate the finger angles for both hands than time to single hand fingers computation. If the directions of both the hands are the same, the computational time will be similar to the single hand computational time. But in real life, it is not always possible that both hands always pointing towards the same direction. Hence, one has to apply this algorithm twice on the image frame to compute for both hands. That will cause extra expense of time and in real time applications it is very much essential that the computational time should be very small. Therefore, a new approach is required for both hand fingers' angle calculation. Figure 1 shows the block diagram flow of our approach.

## II. BACKGROUND WORK

The real time shapes of hand gestures are unknown and could be recognized if the fingers information are correctly known. As the fingertip detection is not possible in bent fingers using color space based techniques, the finger angles need to be computed. The main applications of angle calculation of bent human fingers are in the controlling of machines and robots. Real-time applications need more precise and fast input to machines so that the actuation would be accurate within the given time limit. We couldn't find any previous studies in available literature except [Nolker & Ritter, 2002] that involved calculating the bent fingers' angle without performing any training.

Claudia and Ritter [2002] in her system called 'GREFIT', calculated finger angles with the use of neural network. Chen and Lin [2001] presented a real time parallel segmentation method using mean-C adaptive threshold to detect the region of interest. More applications and segmentation methods for both hands could be found in sign language recognitions and alike systems [Raheja, Chaudhary & Maheswari, 2014][Alon et al., 2009][Jeong, Lee & Kim, 2011]. In our previous work [Chaudhary & Raheja, 2013], a supervised ANN based method has been presented to calculate one hand fingers' angle calculation when fingers are bending in real time. The real time fingertip and centre of palm detection is shown in Figure 2.

Many researchers [Sawah et al., 2007] [Nolker & Ritter, 2002] [Nguyen, Pham, & Jeon, 2009][Lee, & Chun, 2009] [Raheja, Chaudhary, & Singal, 2011] [Gastanldiand et al., 2005] [Kim & Lee, 2008] [Shin, Tsap, & Golgof, 2004][Zhou, & Ruan, 2006] have used fingertip detection as key to detect shape in their research work according based on their applications. Nguyen, Pham and Jeon [2009] presented fingertip detection of both hands. He implemented Claudia's method for hand segmentation. Lee and Chun [2009] used marker-less method in his augmented reality application to register virtual objects where fingertips were detected on curvature of contour.

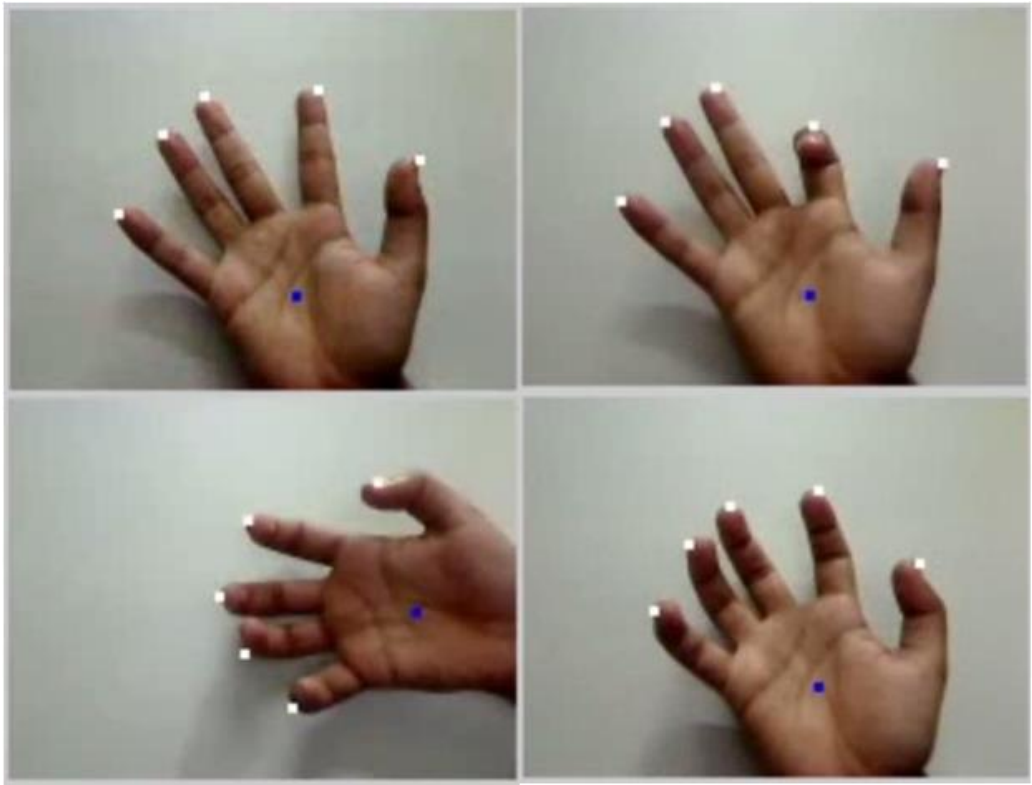

Figure 2 Results of fingertips and centre of palm detection [Chaudhary & Raheja, 2013]

Sawah et al. [2007] showed 3D posture estimation using DBN for dynamic hand gesture recognition, but employed a glove to detect fingers and the palm. Raheja, Chaudhary and Singal [2011] have detected fingertips accurately with the help of MS KINECT® but that doesn't fulfill the natural computing requirements addressed in this paper. Gastaldi et al. [2005] used hand perimeter to select points near fingertips and to reduce the computational time. A detailed review on hand pose estimation is presented by Erol et al. [2007].

### III. FINGERS' ANGLE CALCULATION

In the first phase of this project [Nolker & Ritter, 2002], milestone was to calculate bent finger angles for one hand. The constraints were that the hand can be either right or left as it can't be say in advance that which hand user will use. The developed system works fine with the requirements of real time response. The next phase is to extend the project, if the system need to be controlled using both hands. To calculate both hands finger angles, there are two ways: firstly repeat the single hand method for both hand but it would take double time as for single hand method. Hence, to reduce the computational time, it will be better to process both the hands at the same time. The presented method here takes less processing time since both hands are processed simultaneously. The following text describes region of interest segmentation, fingertips detection, centre of palms detection and bended fingers' angle calculation methods respectively.

### A. Region of Interest Segmentaion

The region of interest (ROI) is needed to extract from original images which makes the work faster and reduces the computational time taken. The HSV color space based skin filter was used to form the binary silhouette of the input image, which will be used to segment hands and masking with original images. The hand segmentation was implemented as described in [Raheja, Das, & Chaudhary, 2011] to obtain the ROI. Essentially, HSV type color spaces are deformations of the RGB color cube and they can be mapped from the RGB space via a nonlinear transformation. The reason behind the selection of this color space in skin detection is that it allows users to intuitively specify the boundary of the skin color class in terms of the hue and saturation. As 'Value' in HSV provides the brightness information, it is often dropped to reduce illumination dependency of skin color. A more robust learning based skin segmentation method is described in [Chaudhary & Gupta, 2012]. The characteristics of skin color with color based segmentation have been discussed in [Raheja et al., 2011].

After the formation of the binary image, BLOBs were collected and the BLOB analysis based on 8 connectivity criteria was applied. As a result of that the two hands were distinguished from each other. They were given different grey values as to ensure that the parameters of one hand are compared against that hand only. Consequently, a mistake, namely consideration of the fingertip of one hand and COP of other hand, is avoided. The hand which seem first in the scene from the left, would be considered as the first hand as there is no concept of right or left hand. The main purpose of BLOB analysis is to extract the two biggest BLOBs to eliminate the false detection of skin pixels and to distinguish the two BLOBs from each other. Figure 3 presents the result of hand segmentation. The brighter BLOB corresponds to the right hand of the main frame and the other BLOB corresponds to the left hand.

simultaneously a new approach based on level set method was developed. The circular separability filter (CSF) and concentric circular filter (CCF) which used level set based numeric values for segmentation, has been taken into consideration [Chaudhary et al., 2012]. The circular separability filter has the shape of a square with a concentric circle inside as shown in Figure 4. After a number of experiments, the radius of the circle is taken to 5 pixels and the bounding square is considered of 20 pixel length. When the filter response is computed for all the points of the region of interest in the binary image, the filter response for the fingertip regions are found distinctively different from that of other regions because of their boundary characteristics. The candidate fingertip locations are determined by using an appropriate threshold condition.

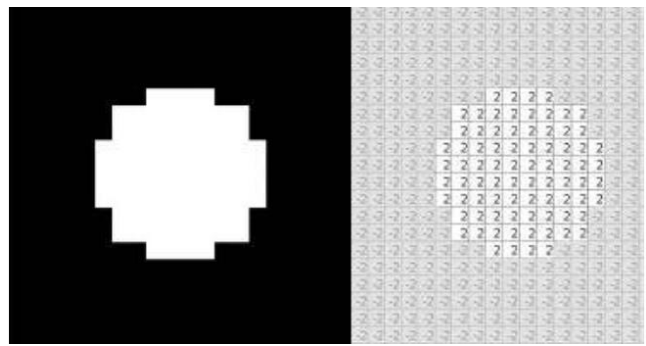

Figure 4: Circular separability filter

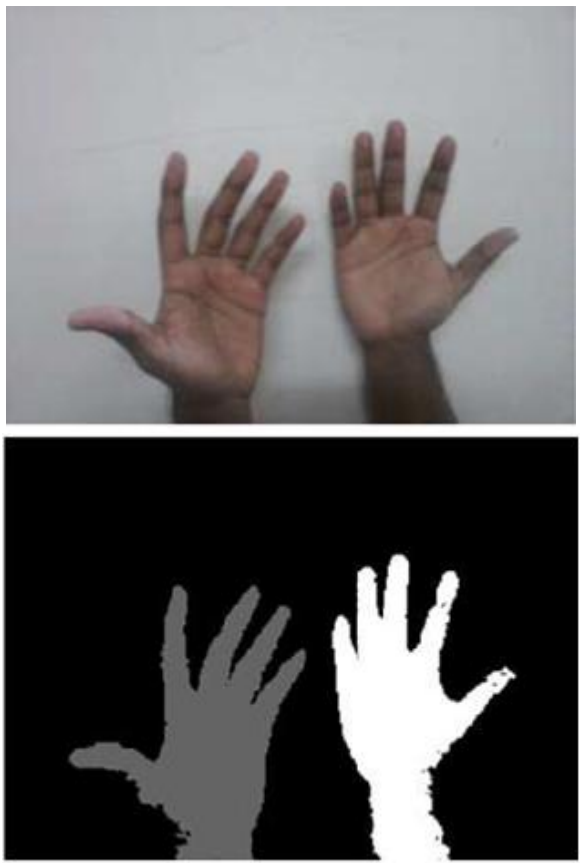

Figure 3: Result of hands segmentation.

After getting the fingertips areas from filter, we need to get exact fingertip location. It is calculated in two steps. Firstly, an approximate location for the fingertip is determined and then using orientation of finger, exact location is calculated. In CSF operation the 8-connected points that satisfy the threshold condition for the filter response of the circular separability filter are grouped together. The groups having a number of pixels greater than a set threshold, are selected and the centroids of the groups are taken as the approximate fingertip positions. Here the first step completes. Then, the orientation of each finger is determined using a filter with 2-concentric circular regions which is again based on numerical values. The working of CCF filter is described in following text. The concentric circular filter is shown in Figure 5.

## B. Fingertip Detection

In the available literature, we couldn't find any study related to the detection of fingertips of both hands in parallel. Although for one hand fingertips has been detected in real time [Chaudhary & Raheja, 2013]. To process both the hands

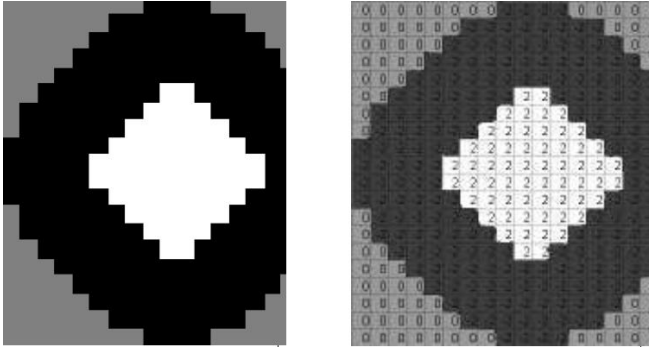

Figure 5:(a) Concentric circular filter and (b) Concentric circular filter element values.

The diameters of the inner and outer circular regions of the CC filter are 10 pixels and 20 pixels respectively. The points inside the inner circle are assigned a value of +2, points in the mid circle are assigned the value of -2, and the points outside the mid circle but inside the bounding square are assigned value of 0, as described in level set method. All these values were set after experimental results. The filter is then applied to the binary silhouette of the hand image which is result of CSF filter. The pixels that lie in the inner circle region are grouped by the 8-connectivity criteria. Then, the largest group is selected and the centroid of the group is calculated.

The orientation of the finger is calculated as the angle ($\theta$) defined by the line joining the centroid of the largest group and the previously calculated approximate finger tip location with the horizontal axis. This process is shown in Figure 6. Then move in this direction several steps with an incremental distance ($r$) using equation (1) and (2) till the edge of the finger is reached.

$$R_{new} = R_{old} + r\cos(-\theta) \qquad (1)$$

$$C_{new} = C_{old} + r\sin(-\theta) \qquad (2)$$

Where $R_{old}$ and $C_{old}$ are the 2D coordinates of the previous trace point, $R_{new}$ and $C_{new}$ are the 2D coordinates of the current trace point and $r$ is the incremental distance. The values of $R_{new}$ and $C_{new}$ after the iterations give the exact coordinates of the fingertips.

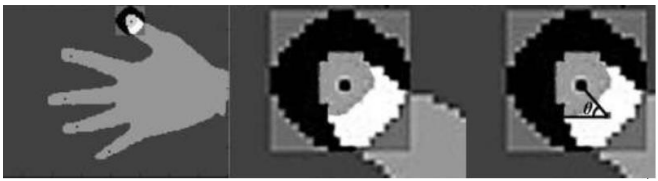

Figure 6:(a) The concentric circular filter being applied on the approximate thumb tip location, (b) Zoomed view of the thumb tip region and (c) Position of the centroid of the largest 8-connected group (region in white) and the angle ($\theta$) with respect to the horizontal.

### C. Centre of Palms Detection

The centre of palm (COP) is needed for further processing. The exact location of the COP in the hand is identified by applying a mask of dimension 30x30 to binary silhouette of the image and counting the number of skin pixels lying within the mask. If the count is within a set threshold, then the centre of the mask will be considered as the candidate for COP. Finally, the mean of all such candidates found in a BLOB are considered as the COP of the hand represented by that BLOB. Since there are two BLOBs, two COPs would be discovered for both the hands. Figure 7 presents the result of fingertip and COP detection. The yellow dots mark the COPs of both hands while the white points are the detected fingertips.

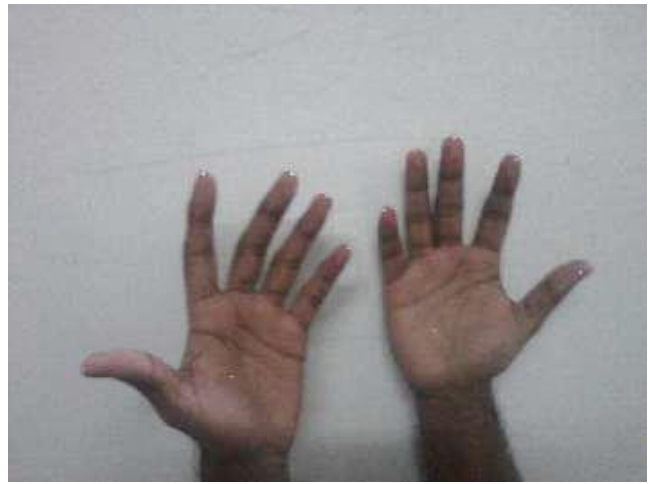

Figure 7: Result of COPs and fingertips detection for both hand.

### D. Angles Calculation for both Hands' Bent Fingers

The presented geometrical method doesn't need any training or sample data to calculate the angles for both hands. The user can use this application with bare hand just showing hands to camera. The palm should face the camera for appropriate operation. The distance between each fingertip and COP can be calculated by subtracting their coordinates on the image frames. Initially the user has to show a reference frame to the system in which all fingers are open and the bending angles of all fingers are $180^0$. The distance between any fingertip and the COP would be the maximum in this position. As the user starts bending the fingers in either direction, distances among fingertips and COP would decrease. The user can move his hands in front of the camera, it is not necessary to have his hand or arm static. This method will calculate the angles in that case also.

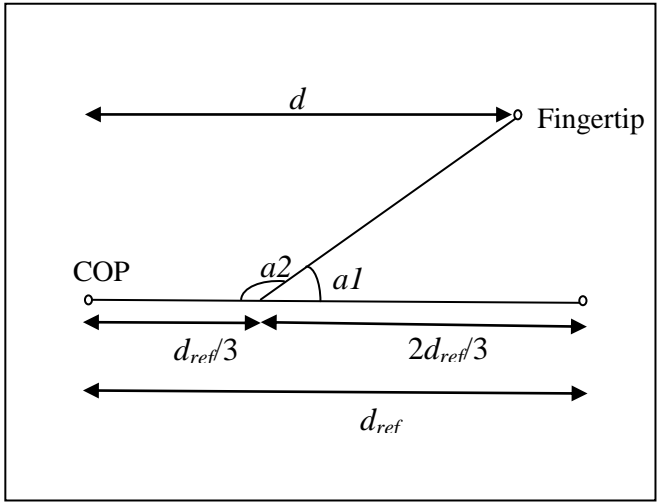

Figure 8: Angle approximation geometry.

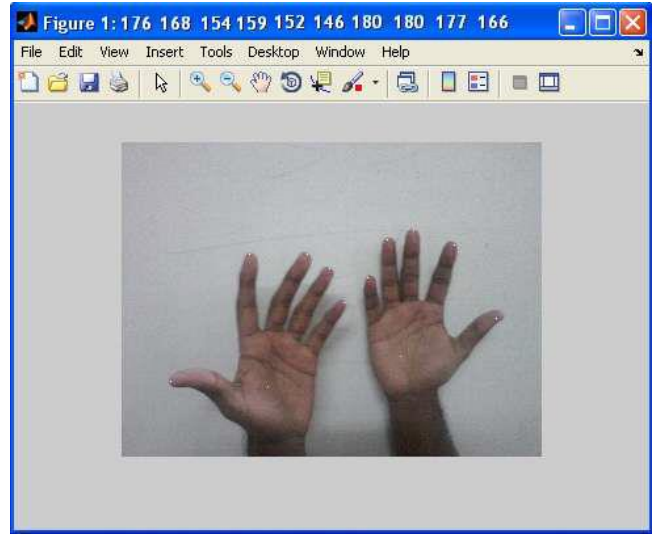

Figure 9: Finger bending angle approximation of double hand.

Here the maximum angle calculated would be $90^0$ as after this limit the fingertips would not be detected using color space based segmentation. The values calculated from the reference frame are stored for each finger. If the user changes the position of his fingers, the distance between COP and fingertips would be compared with the reference distances. The geometrical analysis as described below, calculate the bending angles. From Figure 8 it is clear that

When $d=d_{ref}$, angle $a1=0^0$ and

When $d=d_{ref}/3$, angle $a1=90^0$.

Hence, we can express angle $a1$ as shown in (3).

$$angle\ a1 = 90^0 - \frac{d-\frac{d_{ref}}{3}}{\frac{2d_{ref}}{3}} * 90^0 \qquad (3)$$

Angle $a2$ of finger bending can be obtained from the Figure 7as

$$angle\ a2 = 180^0 - \left(90^0 - \frac{d-\frac{d_{ref}}{3}}{\frac{2d_{ref}}{3}} * 90^0\right) \quad (4)$$

$$or\ \ angle\ a2 = 90^0 + \frac{d-\frac{d_{ref}}{3}}{\frac{2d_{ref}}{3}} * 90^0 \qquad (5)$$

Here the angle $a2$ stores the value of the finger bending angle for one finger. Figure 9 presents the result of fingers angle detection for both hands simultaneously. The angles are shown on the top of window according to the finger shown sequence. This method work in other conditions like a person have only eight fingers in both hands, then the system came to know this automatically and will show only eight angles at display.

## IV. EXPERIMENTAL RESULTS

The experiments were carried out on Intel® i5 processor and 4GB RAM desktop computer. The discussed method is implemented in MATLAB® on Windows® XP. The live video is captured using Logitech® HD webcam with image resolution 240x230. The usage of the system is very similar to what is described in [Chaudhary & Raheja, 2013], only difference is that it will work for both hands simultaneously. If we apply single hand finger's angle calculation method to detect both hands fingers' angle, the computational time come out 294ms. On the other hand, the method proposed in this paper takes only 198ms to perform the same computations.

Both the hands are recognized distinctly, also the system remembers both hands' parameter separately. If there is only one hand shown, system will work perform a single hand gesture analysis. This method provides an accuracy of around 90-92% on live input in varying light conditions. Also the systems take care of hand as well as arm movement and form the analysis in the similar way. It can also inform about the displacement of arm. The numbers of fingers are set maximum as ten but if user have less fingers, it system works fine in that case too.

## V. CONCLUSIONS

This paper presents a novel technique for the bent fingers' angle calculation from the hand gesture in real time. The user has to show bare hands to the system and he is free to bend his fingers. The system would describe bend angles in real time in the same sequence as fingers appeared in the scene from left to right. This technique carries tremendous significance since it can be adopted for identifying human gestures, which are often depicted during communication using both hands. The system considered uses no training data and the hands can be used in any direction.

A. Chaudhary, J.L. Raheja, K. Das, S. Raheja, "Fingers' Angle Calculation using Level-Set Method", Published in Computer Vision and Image Processing in Intelligent Systems and Multimedia Technologies, IGI USA, 2014, pp.191-202.

This approach minimizes the processing time of our last algorithm for determining the angles of a single hand. The processing time is reduced by 96ms, which corresponds to a reduction of approximately 33%. This technique can be used in many applications, one which we have tried is in controlling robotic hands which will mimic hand gestures. In the future we would like to enhance this technique by reducing the time computation and making it more robust on background noise. Also more application would be deployed based on this technique, which will be useful for the public daily life operations.